\documentclass[10pt,twocolumn,letterpaper]{article}

\usepackage[pagenumbers]{cvpr} %

\usepackage{amsmath}
\usepackage{amssymb}
\usepackage{bm}
\usepackage{booktabs}
\usepackage{adjustbox}
\usepackage[dvipsnames]{xcolor}
\usepackage{siunitx}
\usepackage{nicefrac}
\usepackage{multirow}
\usepackage{makecell}
\usepackage{tabularx}
\usepackage{gradient-text}

\newcommand{\nparagraph}[1]{\noindent\textbf{#1.  }}

\usepackage[acronym]{glossaries}
\newacronym{mtmc}{MTMC}{Multi-Target Multi-Camera}
\newacronym{bev}{BEV}{Bird’s Eye View}
\newacronym{moda}{MODA}{Multiple Object Detection Accuracy}
\newacronym{modp}{MODP}{Multiple Object Detection Precision}
\newacronym{iou}{IoU}{Intersection over Union}
\newacronym{fov}{FOV}{Field of View}
\newacronym{mot}{MOT}{Multiple Object Tracking}
\newacronym{mota}{MOTA}{Multiple Object Tracking Accuracy}
\newacronym{motp}{MOTP}{Multiple Object Tracking Precision}
\newacronym{reid}{re-ID}{Re-Identification}
\newacronym{pom}{POM}{probabilistic occupancy map}
\newacronym{nms}{NMS}{non-maximum suppression}
\newacronym{sota}{SOTA}{state-of-the-art}

\usepackage{xspace}

\newcommand{\own}[2]{\makecell[r]{\small{\textbf{#1}} \\[-0.35em] \scriptsize{(#2)}}}

\DeclareUnicodeCharacter{1D54F}{$\mathbb{X}$}

\definecolor{cvprblue}{rgb}{0.21,0.49,0.74}
\usepackage[pagebackref,breaklinks,colorlinks,citecolor=cvprblue]{hyperref}

\def\paperID{4} %
\def\confName{CVPR}
\def\confYear{2024}

\def\sname{TrackTacular\xspace}

\title{Lifting Multi-View Detection and Tracking to the Bird's Eye View}

\author{
Torben Teepe \qquad Philipp Wolters \qquad Johannes Gilg \qquad Fabian Herzog \qquad Gerhard Rigoll\\[0.1cm]
Technical University of Munich
}

\begin{document}
\maketitle
\begin{abstract}
Taking advantage of multi-view aggregation presents a promising solution to tackle challenges such as occlusion and missed detection in multi-object tracking and detection. Recent advancements in multi-view detection and 3D object recognition have significantly improved performance by strategically projecting all views onto the ground plane and conducting detection analysis from a \gls{bev}.
In this paper, we compare modern lifting methods, both parameter-free and parameterized, to multi-view aggregation. 
Additionally, we present an architecture that aggregates the features of multiple times steps to learn robust detection and combines appearance- and motion-based cues for tracking.
Most current tracking approaches either focus on pedestrians or vehicles. In our work, we combine both branches and add new challenges to multi-view detection with cross-scene setups.
Our method generalizes to three public datasets across two domains: (1) pedestrian: Wildtrack and MultiviewX, and (2) roadside perception: Synthehicle, achieving state-of-the-art performance in detection and tracking.
\iftoggle{cvprfinal}{%
\href{https://github.com/tteepe/TrackTacular}{https://github.com/tteepe/\sname}%
}{%
\href{https://anonymous.4open.science/r/TrackTacular-04}{https://anonymous.4open.science/r/\sname-0\paperID}%
}.
\end{abstract}
    
\glsresetall
\section{Introduction}
\label{sec:intro}

\Gls{mtmc} tracking has been a niche topic within the tracking community compared to the more popular \gls{mot} task. Even though using multiple cameras to overcome the challenge of occlusion and missed detections was already introduced in one of the first modern tracking datasets PETS2009 \cite{ferryman2009pets2009}. However, in recent years, the \gls{mtmc} task gained more attention with more cameras beginning to be deployed in the wild and the availability of more \gls{mtmc} datasets \cite{cress2022a9, tang2019cityflow, hou2020multiview, synthehicle2023herzog, gloudemans2023interstate}. In recent years, new approaches were either designed for pedestrian tracking \cite{chavdarova2018wildtrack, hou2020multiview, tang2019cityflow} or for tracking vehicles \cite{synthehicle2023herzog}. In this paper, we want to unify both branches and introduce an approach that can generalize to both domains and outperforms the state-of-the-art on four public datasets: two classical pedestrian datasets: Wildtrack \cite{chavdarova2018wildtrack} and MultiviewX \cite{hou2020multiview} and one vehicle datasets:  Synthehicle \cite{synthehicle2023herzog}. 
The commonly used pedestrian datasets, Wildtrack and MultiviewX, have a few shortcomings for modern computer vision research as they only consist of one scene, and they have a first 90\% of frames train and last 10\% of frames test split, which is prone to overfitting. Thus, we evaluate a new challenging dataset. Synthehicle \cite{synthehicle2023herzog} is a synthetic roadside dataset covering multiple intersection scenarios for vehicle tracking. This dataset requires approaches to generalize over different unseen scenes. 

\begin{figure}
    \centering
    \includegraphics[width=\linewidth]{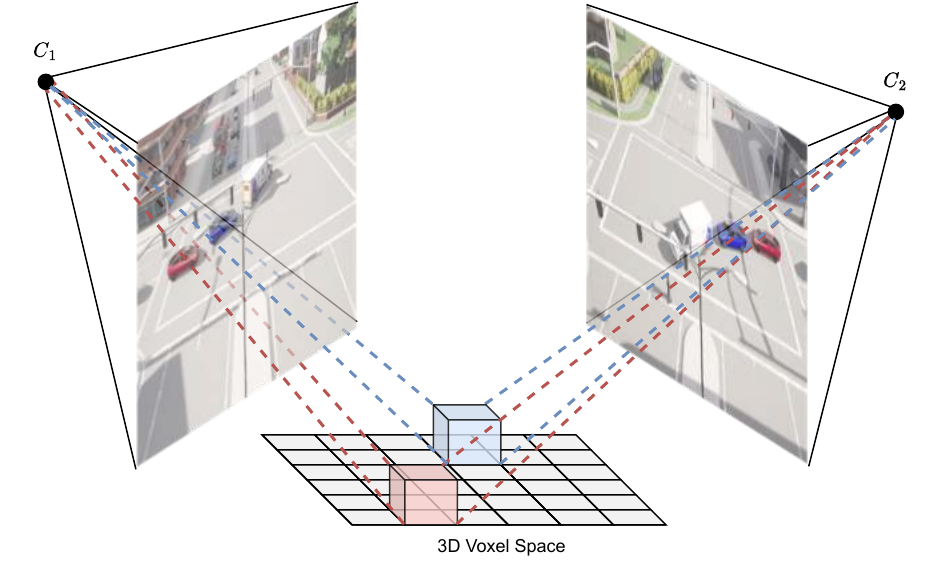}
    \caption{\textbf{Lifting Methods.} We compare three methods that lift the pixel information to 3D voxel \gls{bev} space for detection and tracking.}
    \label{fig:teaser}
\end{figure}

Compared to the single-camera task, the \gls{mtmc} task is more challenging as it requires two association steps: (1) an association between cameras and (2) an association between detections. On the other hand, multi-camera approaches allow for a much stronger 3D perception as the 3D position of the objects can be triangulated from multiple cameras. Traditionally \gls{mtmc} approaches \cite{hofmann2013hypergraphs, synthehicle2023herzog, nguyen2022lmgp} start with 2D bounding box detection and then create camera tracklets that are associated to form a global track. Other approaches used 2D detection to predict a 3D location and perform the association in 3D \cite{cheng2023rest}. More recently, approaches started skipping the 2D detection step; they project the features to the \gls{bev} and perform both detection and tracking in 3D \cite{teepe2023earlybird}. These works show that fusing the data earlier in 3D yields better results than late-fusion approaches. We want to apply this early-fusion mindset and propose a new architecture that uses more advanced methods to project the features in a 3D space.

The projection of camera features has been studied extensively in the perception models \cite{li2022bevformer, harley2022simple, philion2020lift} for autonomous vehicles and has become an essential part of the multi-sensor perception system. While other sensors like Lidar and Radar have depth information, the camera image is a projection of the 3D world onto a 2D plane, with no trivial way to reverse this projection. Thus, lifting methods have been developed to recover 3D information. The projection is also essential to fuse camera data with other sensors \cite{wolters2024unleashing}.

Nevertheless, there are critical differences between vehicle perception and multi-view detection:
(1) The cameras in our approach have a larger overlapping area and thus can aggregate information from multiple cameras. (2) Our cameras are mostly static; thus, we can exploit the scene's geometry (3) The observed area is much larger, and the cameras are further away from the targets. However, the overlap gives us a key advantage: We can triangulate the 3D position of the targets using an approximation to good ol' epipolar geometry \cite{hartley2003multiple}.

\begin{figure*}[!htb]
    \centering
    \begin{subfigure}[b]{0.325\textwidth}
      \includegraphics[width=\textwidth]{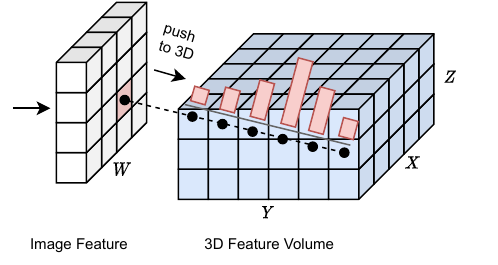}
      \caption{In the depth splatting approach \cite{philion2020lift} the 2D feature are \textit{pushed} to 3D with depth prediction, filling voxels that intersect with its ray.}
      \label{fig:depth_lift}
    \end{subfigure}
    \hfill
    \begin{subfigure}[b]{0.325\textwidth}
      \includegraphics[width=\textwidth]{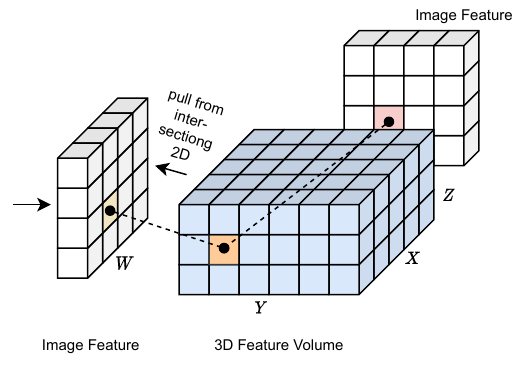}
      \caption{The bilinear sampling method introduced in Simple-BEV \cite{harley2022simple} each 3D Voxel \textit{pulls} feature from the 2D map by projection and sampling.}
      \label{fig:simple_bev}
    \end{subfigure}
    \hfill
    \begin{subfigure}[b]{0.325\textwidth}
      \includegraphics[width=\textwidth]{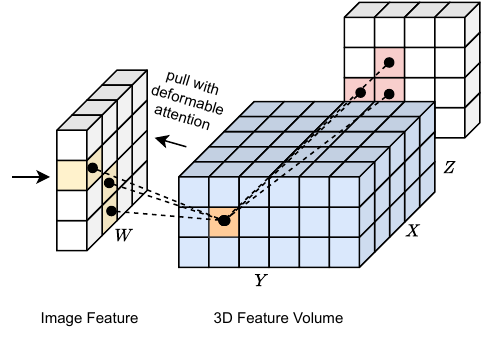}
      \caption{The lifting method introduced by BEVFormer \cite{li2022bevformer} uses deformable attention to aggregate multiple image features with a learnable offset.}
      \label{fig:lift_triangulation}
    \end{subfigure}
    \caption{\textbf{Lifiting Methods.} The three lifting methods we compare in this paper. The bilinear sampling method (b) simplifies the depth splat approach (a) without explicitly predicting the depth. Our method extends the bilinear sampling to only project image features if they intersect in the 3D volume. Thus, our method approximates the triangulation at voxel granularity.}
    \label{fig:lifitng}
  \end{figure*}

Tracking is also part of autonomous perception; the challenge is more manageable than tracking with static cameras, as the targets are within a small surrounding area and are only observed shortly. Thus, most methods \cite{wang2023exploring, yin2021center} use simple distance-based association methods. 
In our approach, we face more significant scenes with more targets that need to be tracked over extended periods. Staying with the early-fusion approach, we propose a novel association method that uses the history \gls{bev} information to predict the location of each detection in the previous frame. It combines the advantages of appearance-based and motion-based association and learns to combine both cues. 

\noindent In summary, our contributions are as follows:
\begin{itemize}
    \item We combine our novel tracking strategy with three existing lifting methods and extend them for views with strong overlap to show state-of-the-art detection and tracking results on three public datasets.
    \item We propose a novel learned association method that combines the advantages of appearance-based and motion-based association and outperforms previous methods in tracking.
    \item We set a new strong baseline on more challenging datasets for \gls{mtmc} with a standard evaluation protocol to initiate further and more comparable research in this field.
\end{itemize}

\section{Related Work}

\nparagraph{Multi-View Object Detection}
Multi-camera systems are widely used to tackle the challenge of pedestrian detection in highly occluded environments. Such systems comprise synchronized and calibrated cameras that capture a common area from various angles. The multi-view detection system then processes the overlapping views to detect pedestrians.
MVDet \cite{hou2020multiview} introduced an approach that uses convolutions to train models end-to-end, projecting encoded image features from each perspective onto a shared ground plane. This process led to noticeable improvements and has been the basis for subsequent methods, including ours. Projections are not just limited to the sparse detections from each viewpoint. The method introduced by \cite{hou2020multiview} encodes the input image and projects all features onto the ground plane using a perspective transformation. This mapping results in distortions on the ground plane like a shadow of the actual object \cite{hou2021multiview}.
To overcome the  limitations of perspective transformation, several other methods \cite{hou2021multiview, lee2023multi, song2021stacked} have been proposed. One approach \cite{hou2021multiview} uses projection-aware transformers with deformable attention in the \gls{bev}-space to move the "shadows" back to their original location. Another method \cite{lee2023multi} uses regions of interest from 2D detections and separately projects these to the estimated foot position on the ground plane. A third approach \cite{song2021stacked} uses multiple stacked homographies at various heights to better approximate a complete 3D projection.
Shifting the focus from model improvement, \cite{qiu20223d} attempted to enhance detection by improving the data. This approach added more occlusions by introducing 3D cylindrical objects. Such a data augmentation reduces the dependence on multiple cameras, helping to avoid overfitting.

Overall, our approach follows the tracks of MVDet \cite{hou2020multiview}, and we include their projection method as our baseline. However, among other improvements, we will also explore other projection methods explained in the next paragraph to improve the detection performance.

\nparagraph{3D Perception Systems}
Multi-sensor perception systems are mainly developed for autonomous vehicles to fuse the data from different sensors. With the enormous interest in autonomous driving, this area is progressing rapidly. For this section, we want to focus on approaches that focus on camera lifting.

The first and simplest unprojection are homography-based methods \cite{hou2020multiview, hou2021multiview, lee2023multi, song2021stacked}. They assume a flat ground plane and use a homography matrix to project the image features to the ground plane. While this method is parameter-free, it is less accurate for objects above the ground plane and causes shadow-like artifacts for objects far away from the camera \cite{hou2021multiview}. Simple-BEV \cite{harley2022simple} proposed another parameter-free projection method: it defines a 3D volume of coordinates over the BEV plane, projects these coordinates into all images, and averages the features sampled from the projected locations \cite{harley2019learning}. Compared to homography-based projection, this method \textit{pulls} information from the image to a 3D location instead of \textit{pushing} information from the image to the world. Depth-based approaches \cite{philion2020lift} employ a monocular depth estimator that estimates a per-pixel depth to project the pixels into the 3D space. This method is very effective as it does not require depth information to be explicitly available and can also deduct it from the scene. Another way of lifting the image features is BEVFormer \cite{li2022bevformer}. It is similar to Simple-BEV \cite{harley2022simple} as it aggregates from all images to a 3D location. However, it uses a deformable attention mechanism to learn the aggregation weights. This method is more flexible than Simple-BEV \cite{harley2022simple} as it can learn to focus on specific objects, but it comes with a much higher computational cost.
In our work, we will compare these different lifting strategies and extend them to explicitly enforce the triangulation of features.

\nparagraph{Multi-Target Multi-Camera Tracking} 
There is a wealth of research on single-camera tracking, which will be discussed later. However, we concentrate on \gls{mtmc} tracking in this section. Most \gls{mtmc} trackers base their models on the assumption of an overlapping \gls{fov} among cameras. The method by Fleuret et al. \cite{fleuret2007multicamera} makes use of this overlapping \gls{fov} to represent targets within a \gls{pom}, combining color and motion features with occupancy probabilities during the tracking process. \cite{berclaz2011multiple} enhanced this approach by framing tracking within \gls{pom}s as an integer programming problem, solving it optimally using the k-shortest paths (KSP) algorithm.

\gls{mtmc} tracking can also be interpreted as a graph problem. Hypergraphs \cite{hofmann2013hypergraphs} or multi-commodity network flows \cite{shitrit2013multi, leal2012branch} are used to establish correspondences across views, which are then resolved using min-cost \cite{hofmann2013hypergraphs, shitrit2013multi} or branch-and-price algorithms \cite{leal2012branch}.

\begin{figure*}[th!]
    \centering
     \includegraphics[width=.95\linewidth]{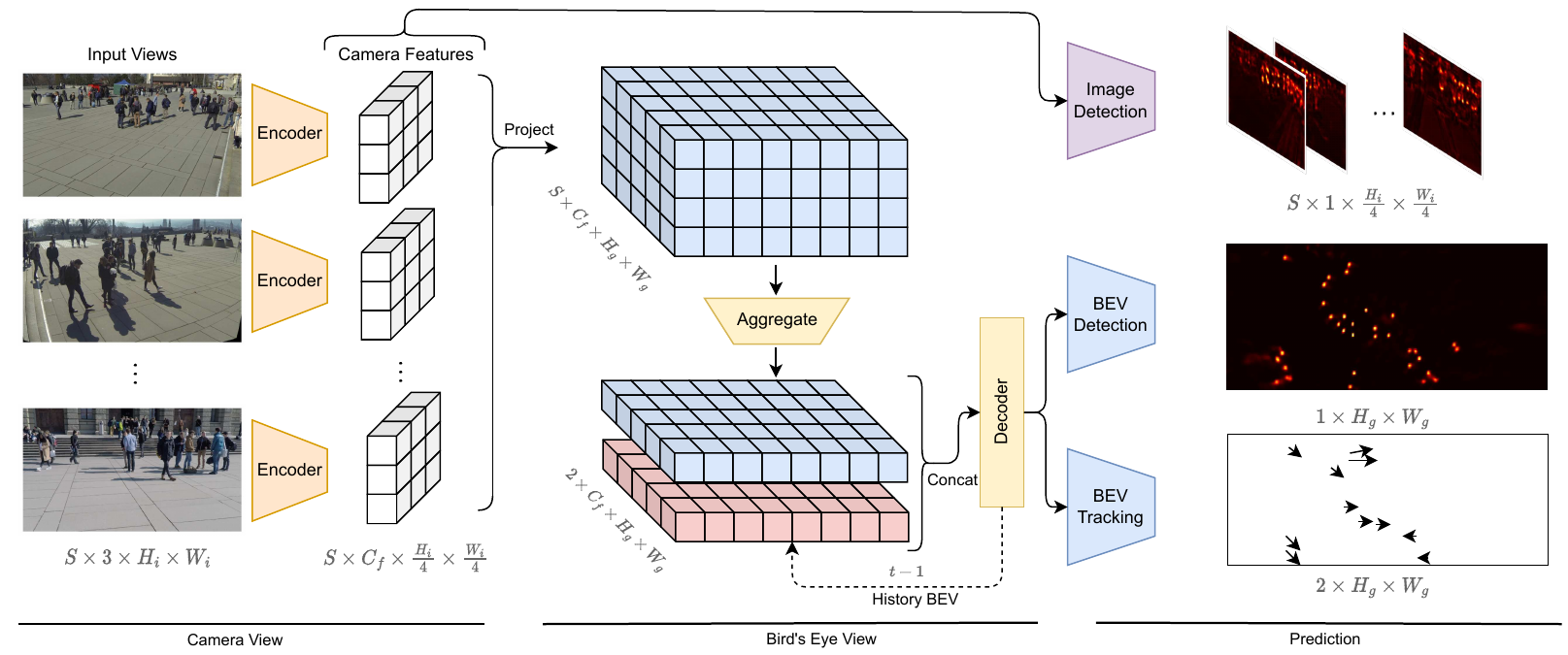}
  
     \caption{\textbf{Overview of Our Approach.} The input views are encoded, and the resulting camera features are projected using one of three lifting methods. After aggregation, the \gls{bev} feature is concatenated with the feature of the previous step. With the decoded \gls{bev} feature, we predict the locations and offset to the location in the previous step. Additionally, we guide the architecture by predicting the object centers in the image features.}
     \label{fig:overview}
  \end{figure*}

Recently, a two-stage approach has gained popularity: it begins with generating local tracklets for all targets within each camera, followed by matching local tracklets corresponding to the same target across all cameras. The task of generating local tracklets within a single camera, known as single camera \gls{mot}, has been extensively studied \cite{bergmann2019tracking, zhang2021fairmot, zhou2020tracking, wang2020towards, feichtenhofer2017detect, Wojke2017simple, chen2018real}. With the advancement in object detection techniques, tracking-by-detection \cite{feichtenhofer2017detect, seidenschwarz2023simple,bergmann2019tracking, zhou2020tracking, Wojke2017simple} is now the favored method for multi-target tracking. For the second step, several strategies for cross-view data association have been proposed to match local tracklets across different cameras. Some studies \cite{hu2006principal, eshel2008homography} use the principles of epipolar geometry to find correspondences based on location on the ground plane. Besides ground plane locations, \cite{xu2016multi} incorporates appearance features as cues for the association.
Cutting-edge models \cite{nguyen2022lmgp, cheng2023rest} have inverted the initial two steps: the 2D detections are first projected onto the 3D ground plane, and a graph is created using \gls{reid} node features. These nodes are either assigned spatially and temporally at the outset \cite{cheng2023rest} or both assignments are made simultaneously \cite{nguyen2022lmgp} utilizing graph neural networks for link prediction. While all current methods \cite{cheng2023rest, seidenschwarz2023simple,bergmann2019tracking, zhang2021fairmot, zhou2020tracking} are evaluated based on detection results to account for inaccuracies in detection, LMGP \cite{nguyen2022lmgp} uses ground truth bounding boxes and thus cannot be compared to recent studies.

Our approach stands apart from all previous work and aligns more closely with one-shot trackers, which are discussed in the subsequent section. Similar to the latest methods \cite{nguyen2022lmgp, cheng2023rest}, our approach establishes spatial associations within our detector, then proceeds to associate on the ground plane.

\nparagraph{One-Shot Tracking} 
A subset of single-view Multi-Object Trackers includes one-shot trackers. These systems conduct detection and tracking in a single step, saving computation time; however, they generally perform worse than two-step trackers. The features predicted can either be \gls{reid} features \cite{zhang2021fairmot, Voigtlaender19mots, wang2020towards} or motion cues \cite{zhou2020tracking, bergmann2019tracking, feichtenhofer2017detect}.

Track-RCNN \cite{Voigtlaender19mots} is the first example of a \gls{reid}-based approach. It adds \gls{reid} feature extraction to Mask R-CNN \cite{he2017mask}, generating a bounding box and a corresponding re-ID feature for each proposal. Similarly, JDE \cite{wang2020towards} is built upon YOLOv3 s\cite{redmon2018yolov3}, and FairMOT is based on CenterNet \cite{zhou2019objects}. The key advantage of FairMOT, compared to the others, is its anchor-free design, meaning that detections do not rely on a fixed set of anchor bounding boxes but on a singular detection point, enhancing the separation of \gls{reid} features.

Motion-based trackers such as D\&T \cite{feichtenhofer2017detect} take input from adjacent frames, predicting inter-frame offsets between bounding boxes. Tracktor \cite{bergmann2019tracking} exploits the bounding box regression head to associate identities, thus removing box association. In contrast, CenterTrack \cite{zhou2020tracking} predicts the object center offset using a triplet input, consisting of the current frame, the previous frame, and the heatmap of the last frame detection. This prior heatmap allows objects to be matched anywhere, even in the case of overlapping boxes. However, these motion-based methods, which only associate objects in adjacent frames without re-initializing lost tracks, can struggle in managing occlusions.

In our approach, we propose to learn the association between detections in the previous and current time steps, as CenterTrack \cite{zhou2020tracking} and D\&T \cite{feichtenhofer2017detect} proposed. However, we apply this idea at \gls{bev} feature level. This approach combines the advantages of appearance-based and motion-based association and learns to combine both cues. Combined with a Kalman Filter \cite{kalman1960new}, we can also re-initialize lost tracks.

\section{Methodology}

We provide an architecture overview in \cref{fig:overview}. It starts with the $S$ RGB input images ($S \times 3 \times H_i \times W_i$) that are augmented and fed to the encoder network to yield our downsampled image features ($S \times C_f \times \frac{H_i}{4} \times \frac{W_i}{4}$). With different projection methods, features are then projected to a common \gls{bev} space ($S \times C_f \times H_g \times W_g$). In the following step, the \gls{bev} space is then reduced in the vertical dimension ($S \times H_g \times W_g$). The feature of the previous time step is subsequently concatenated to the current \gls{bev} feature ($2 \times S \times C_f \times H_g \times W_g$). The \gls{bev} features are finally fed through a decoder network. 

\subsection{Lifting Methods}

The projection is central to this approach as it provides the link between the image view and the \gls{bev}-view. 

\nparagraph{Perspective Transformation} This method is the simplest lifting method as it does not model height information. Following \cite{hou2020multiview}, we employ perspective projection to transfer the image features onto the ground plane. The pinhole camera model \cite{hartley2003multiple} uses a $3\times4$ transformation matrix $\bm{P} = \bm{K} \left[ \bm{R}|\bm{t}\right]$ that maps 3D locations $(x, y, z)$ to 2D image pixel coordinates $(u, v)$. We choose to project all pixels to the ground plane at $z=0$, which simplifies the projection to:
\vspace{-0.5em}\begin{equation}
    \label{eq:perspective_3x3}
    \resizebox{0.78\linewidth}{!}{$
        d\begin{pmatrix}u \\ v \\ 1\end{pmatrix} = \bm{P_0} \begin{pmatrix} x \\ y \\ 1\end{pmatrix} = \begin{bmatrix} p_{11} & p_{12} & p_{14} \\ p_{21} & p_{22} & p_{24} \\ p_{31} & p_{32} & p_{34} \end{bmatrix} \begin{pmatrix} x \\ y \\ 1\end{pmatrix},
    $}\end{equation}
where $s$ is a real-valued scaling factor and $\bm{P_0}$ denotes the $3\times3$ perspective transformation matrix without the third column from $\bm{P}$. 
Features from all $S$ cameras are projected to the ground plane using this equation, with each having its unique projection matrix $\bm{P_0^{(s)}}$.

\nparagraph{Depth Splatting} The Depth Splat method \cite{philion2020lift} is based on monocular depth estimation. The idea is to simulate a point cloud from a camera image. Each image pixel $(u, v)$ is associated with a discrete depth $d \in D = \{d_0 + \Delta, ..., d_0 + |D| \Delta\}$. This depth distribution is predicted as part of the image features, making this a parameterized lifting method. Unprojecting the image feature channels along the predicted depth yields our point cloud of ($D \times C_f \times \frac{H_i}{4} \times \frac{W_i}{4}$) in each camera frustum. The point clouds are then wrapped into a common voxel space, weighting the channel information with the probably of the discrete depth (\cf \cref{fig:depth_lift}). This method is considered to \textit{push} its information from 2D to 3D.

\nparagraph{Bilinear Sampling} The idea of Simple-BEV \cite{harley2022simple} is to simplify the method of depth projection without explicitly predicting the depth. Each ray would fill its information into all voxels that intersect with it. However, the projection is turned around for this method: the 3D voxels \textit{pull} the information from the 2D image. This pulling is done by projecting the 3D voxel to the image plane, determining if the projected point is inside the image, and later sampling the image features sub-pixel accurate to the voxel. We can project all eight vertices of a voxel to all image planes with
\begin{equation}
    \label{eq:backprojection}
        d\begin{pmatrix}u_n \\ v_n \\ 1\end{pmatrix} = \bm{K} \left[ \bm{R}|\bm{t}\right] \begin{pmatrix} x_n \\ y_n \\ z_n\\ 1\end{pmatrix} = \bm{P^{(s)}} \begin{pmatrix} x_n \\ y_n \\ z_n\\ 1\end{pmatrix},
\end{equation}
and sample the image feature from $[\min(u_n), \min(v_n), \max(u_n), \max(u_n)]$. The features from all $S$ cameras are then averaged for each voxel.
The advantage is that every voxel will receive a feature, while in splatting methods, some voxels further away from the cameras might not be filled at all. This property makes this method more robust in long-range perception. Even though the bilinear sampling method \cite{harley2022simple, harley2019learning} was not designed for multi-camera perception, it reveals a unique property in the overlapping areas: it approximates a triangulation of image points at voxel granularity. This triangulation is illustrated in \cref{fig:lift_triangulation}. While for data with complete overlap, i.e., every voxel is at least seen by two cameras, the bilinear sampling method is equivalent to the triangulation method.

\nparagraph{Deformable Attention} The lifting method introduced in BEVFormer \cite{li2022bevformer} uses an approach that uses each voxel as a query and projects the 3D reference points back to the 2D image views with the \cref{eq:backprojection}. The 2D reference points for each query and features around those image feature locations are sampled. Finally, the features are aggregated as a weighted sum as the output of spatial cross-attention. The approach is similar to bilinear sampling, but the aggregation uses surrounding features of the projected location and aggregates the BEV features with attention.

Instead of the temporal self-attention introduced in BEVFormer, we use the same resource-efficient temporal aggregation introduced in the next section for all lifting methods.

\subsection{Temporal Aggregation}
The core of tracking is to aggregate temporal information. In our architecture, we want to fuse these features early instead of at the detection level. Temporal information can also improve the detection, as detections can not disappear between time steps. The availability of the previous features enables the architecture to learn the motion of each detection. Trackers are usually divided into appearance-based and motion-based; however, our feature contains both types of information, and our architecture can fuse both cues at the feature level. We implement the temporal aggregation in a \textit{late-to-early} fusion \cite{he2023lef} manner: the decoded \gls{bev} feature of the previous timestep is concatenated to the current, undecoded feature (\cf \cref{fig:overview}).

\subsection{Detection \& Tracking Heads}
The general head architectures follow CenterNet \cite{zhou2019objects}, and our main detection branch predicts a heatmap or \gls{pom} on the ground plane. We add another head for offset prediction $(x,y)$ that helps predict the location more accurately as it mitigates the quantization error from the voxel grid. We train the center head with Focal Loss \cite{lin2017focal}, and the offset head with L1 Loss. We also add a detection head to the image features that predict the center of the 2D bounding boxes, helping to guide the features before we project them to voxel space.

For tracking, we predict the motion of each detection in the \gls{bev} space. Similar to the offset, we learn the offset to the location in the previous frame. As these offsets can vary in magnitude, we choose the Smooth L1 Loss.

\begin{table*}[th]
    \centering
    \setlength{\tabcolsep}{4pt}
    \resizebox{.835\linewidth}{!}{%
    \begin{tabular}{rlcccccccc}
    \toprule
     &\multirow[b]{2}{*}{\small{\makecell[l]{Lifting\\ Method}}} & \multicolumn{4}{c}{Wildtrack} & \multicolumn{4}{c}{MultiviewX} \\\cmidrule(lr){3-6}\cmidrule(lr){7-10}
     && MODA & MODP  & Precision  & Recall & MODA  & MODP  & Precision  & Recall \\\midrule
    DeepMCD \cite{chavdarova2017deep}     & \small{Learned} & 67.8  & 64.2  & 85    & 82     & 70.0  & 73.0  & 85.7  & 83.3 \\ 
    Deep-Occlusion \cite{baque2017deep}   & \small{Learned}& 74.1  & 53.8  & 95    & 80     & 75.2  & 54.7  & 97.8  & 80.2 \\ 
    MVDet \cite{hou2020multiview}         & \small{Persp. Proj.} & 88.2  & 75.7  & 94.7  & 93.6   & 83.9  & 79.6  & 96.8  & 86.7 \\
    SHOT \cite{song2021stacked}           & \small{Persp. Proj.} & 90.2  & 76.5  & 96.1  & 94.0   & 88.3  & 82.0  & 96.6  & 91.5 \\
    3DROM$^\dagger$ \cite{qiu20223d}      & \small{Persp. Proj.} & 91.2  & 76.9  & 95.9  & 95.3   & 90.0  & 83.7  & 97.5  & 92.4 \\
    MVDeTr  \cite{hou2021multiview}       & \small{Persp. Proj.} & 91.5  & \textbf{82.1}  & 97.4  & 94.0   & 93.7  & 91.3  & \textbf{99.5}  & 94.2 \\
    EarlyBird \cite{teepe2023earlybird}   & \small{Persp. Proj.} & 91.2  & 81.8  & 94.9  & 96.3   & 94.2  & 90.1  & 98.6  & 95.7 \\
    MVTT \cite{lee2023multi}              & \small{Persp.+RoI} & \textbf{94.1}  & 81.3  & \textbf{97.6}  & \textbf{96.5}   & 95.0  & \textbf{92.8}  & 99.4  & 95.6 \\
    \midrule
    \multirow{4}{*}{\textbf{\sname}} & \small{Persp. Proj.} & 91.8 & 79.8 & 96.2 & 95.6 & 95.9 & 89.2 & \textbf{99.5} & 96.4  \\
     & \small{Bilin. Sampl.}  & 92.1 & 76.2 & 97.0 & 95.1 & \textbf{96.5} & 75.0 & 99.4 & \textbf{97.1} \\
     & \small{Depth Splat.}  & 93.2 & 77.5 & 97.3 & 95.8 & 96.1 & 90.4 & 99.0 & \textbf{97.1} \\
     & \small{Deform. Attn.}  & 78.4 & 73.1 & 93.8 & 84.0 & 94.4 & 73.1 & 98.6 & 95.8 \\
    \bottomrule
    \end{tabular}}
    \caption{\textbf{Pedestrian Detection.} Comparison of the \gls{bev} detection performance with the state-of-the-art methods on the Wildtrack and MultiviewX datasets.  $^\dagger$ 3DROM results are without additional data augmentations.}
    \label{tab:detection-sota}
    \end{table*}

\section{Experiments}\label{sec:experiments}

\subsection{Datasets}

\noindent\textbf{Wildtrack} \cite{chavdarova2018wildtrack} comprises real-world footage obtained from seven synchronized and calibrated cameras. These cameras capture an overlapping field-of-view of a 12$\times$36 meter public area where pedestrian movement is unscripted. Ground plane annotations are offered on a 480$\times$1440 grid, equating to 2.5 cm grid cells. On average, each frame contains 20 pedestrians covered by 3.74 cameras. The video is recorded at a resolution of 1080$\times$1920 pixels at a frame rate of 2 fps.

\noindent\textbf{MultiviewX} \cite{hou2020multiview} is a synthetic dataset modeled close to the specification of Wildtrack dataset using a game engine. This dataset includes views from 6 virtual cameras with an overlapping field-of-view encompassing a slightly smaller area (16$\times$25 meters compared to 12$\times$36 meters in Wildtrack). Annotation are provided on a ground plane grid of size 640$\times$1000, with each grid representing the same 2.5 cm squares. With an average of 40 pedestrians per frame and coverage of 4.41 cameras per location, camera resolution and frame rate are the same as in the Wildtrack dataset. Like Wildtrack, the dataset has a length of 400 frames.

\begin{table}[t]
    \renewcommand\cellgape{\Gape[-1pt][-1pt]}
    \setlength{\tabcolsep}{2pt}
    \centering
    \resizebox{\linewidth}{!}{%
    \begin{tabular}{rcccccc}
    \toprule
    & \multicolumn{5}{c}{Wildtrack}\\\cmidrule(lr){2-6}
    & IDF1$\uparrow$ & MOTA$\uparrow$ & MOTP$\uparrow$ & MT$\uparrow$ & ML$\downarrow$ \\
    \midrule
    \small{KSP-DO~\cite{chavdarova2018wildtrack}}        & 73.2 & 69.6 & 61.5 & 28.7 & 25.1\\ 
    \small{KSP-DO-ptrack~\cite{chavdarova2018wildtrack}} & 78.4 & 72.2 & 60.3 & 42.1 & 14.6\\ 
    \small{GLMB-YOLOv3~\cite{ong2020bayesian}}           & 74.3 & 69.7 & 73.2 & 79.5 & 21.6\\ 
    \small{GLMB-DO     \cite{ong2020bayesian}}           & 72.5 & 70.1 & 63.1 & \textbf{93.6} & 22.8\\ 
    \small{DMCT  \cite{you2020real}}                     & 77.8 & 72.8 & 79.1 & 61.0 & 4.9\\ 
    \small{DMCT Stack  \cite{you2020real}}               & 81.9 & 74.6 & 78.9 & 65.9 & 4.9\\
    \small{ReST$^\dagger$  \cite{cheng2023rest}}         & 86.7 & 84.9 & 84.1 & 87.8 & 4.9\\
    \small{EarlyBird \cite{teepe2023earlybird}}          & 92.3 & 89.5 & \textbf{86.6} & 78.0 & 4.9\\
    \small{MVFlow \cite{engilber2023multi}}              & 93.5 & 91.3 & $-$  & $-$  & $-$ \\
    \midrule
    \own{\sname}{Perspective Transform} & 94.2 & 89.6 & 81.7 & 87.8 & 4.9\\
    \own{\sname}{Bilinear Sampling}     & \textbf{95.3} & \textbf{91.8} & 85.4 & 87.8 & 4.9\\  
    \own{\sname}{Depth Splatting}       & 93.6 & 90.2 & 84.2 & 82.9 & 4.9\\
    \own{\sname}{Deformable Attention}  & 88.0 & 82.2 & 78.9 & 75.6 & 4.9\\ 
    \midrule
    & \multicolumn{5}{c}{MultiviewX}\\\cmidrule(lr){2-6}
    & IDF1$\uparrow$ & MOTA$\uparrow$ & MOTP$\uparrow$ & MT$\uparrow$   & ML$\downarrow$ \\
    \midrule
    \small{EarlyBird \cite{teepe2023earlybird}}  & 82.4 & 88.4 & 86.2 & 82.9 & \textbf{1.3} \\
    \midrule
    \own{\sname}{Perspective Transform} & 84.2 & 91.4 & \textbf{86.7} & 85.5 & 2.6 \\
    \own{\sname}{Bilinear Sampling}     & \textbf{85.6} & \textbf{92.4} & 80.1 & \textbf{92.1} & 2.6\\
    \own{\sname}{Depth Splatting}       & 83.4 & 91.8 & 84.7 & 90.8 & 2.6\\
    \own{\sname}{Deformable Attention}  & 84.8 & 91.4 & 80.6 & 89.5 & 2.6\\ 
    \bottomrule
    \end{tabular}}
    \caption{\textbf{Pedestrian Tracking.} Evaluation of tracking results on the Wildtrack and MultiviewX.
    $^\dagger$Re-computed by us.}
    \label{tab:track-sota}
\end{table}

\noindent\textbf{Synthehicle}~\cite{synthehicle2023herzog} is a synthetic dataset modeling intersections cameras for intelligent cities in CARLA \cite{dosovitskiy2017carla}. 3-8 cameras record each intersection with a large overlapping area in the center of the intersection. The dataset models day, dawn, night, and rain conditions. The scenes are annotated per camera with a camera calibration. We consider the classes cars, trucks, and motorbikes. The dataset has separate towns for the test set, which allows for evaluating unseen intersections.

\begin{table*}[!hbt]
    \centering
    \begin{tabular}{clcccccccc}
    \toprule
    && \multicolumn{3}{c}{Scene Specific} & \multicolumn{3}{c}{Cross Scene}\\\cmidrule(lr){3-5}\cmidrule(lr){6-8}
    Synthehicle & \small{Lifting Method} &  IDF1$ \uparrow$ & MOTA$\uparrow$ & MOTP$\uparrow$ &  IDF1$ \uparrow$ & MOTA$\uparrow$ & MOTP$\uparrow$  \\
    \midrule
    \multirow{2}{*}{\textbf{\sname}} &\small{Bilinear Sampling} & 48.0 & 10.6 & 32.8 & 18.3 & -22.0 & 29.3 \\
    & \small{Depth Splatting} & 57.2 & 33.1 & 41.9 & 24.2 & -1.5 & 32.4 \\
    \bottomrule
    \end{tabular}
    \caption{\textbf{Synthehicle.} Evaluation on the scene specific validation set and the cross scene test set. The validation set consists of temporally separated scenes from the train set and the test set contains only unseen scenes.}
    \label{tab:synthehicle}
\end{table*}

\subsection{Evaluation Metrics}
There are different philosophies on how to evaluate 3D detection. Three common protocols are used: 3D bounding boxes \cite{weng20203d, geiger2013vision}, 2D \gls{bev} bounding boxes\cite{geiger2013vision}, 2D \gls{bev} center points \cite{caesar2020nuscenes}.
We follow the 2D \gls{bev} center point protocol as it is the most common protocol for \gls{mtmc} tracking, and it is more forgiving of minor errors in 3D detection.

\nparagraph{Detection}
Pedestrian detection is classified as true positive if it is within a distance $r = \qty{0.5}{\ meter}$, which roughly corresponds to the radius of a human body. Following previous works \cite{chavdarova2018wildtrack, hou2020multiview}, we use \gls{moda} as the primary performance indicator, as it accounts for the normalized missed detections and false positives. Additionally, we report the \gls{modp}, Precision, and Recall.

\nparagraph{Tracking}
Aligning with our detection philosophy, we evaluate all tracking metrics on 2D \gls{bev} center points \cite{caesar2020nuscenes, chavdarova2018wildtrack}. We report the common \gls{mot} metrics \cite{bernardin2008evaluating, weng20203d} and identity-aware metrics \cite{ristani2016performance}. The threshold for a positive assignment is set to $r = \qty{1}{\ meter}$. The primary metrics under consideration are \gls{mota} and IDF1.
\gls{mota} takes missed detections, false detections, and identity switches into account. IDF1 measures missed detections, false positives, and identity switches.

\subsection{Implementation Details}
Following \cite{harley2022simple, hou2021multiview}, we apply random resizing and cropping on the RGB input in a scale range of $[0.8, 1.2]$ and adapt the camera intrinsic $K$ accordingly. Additionally, we add some noise to the translation vector $\bm{t}$ of the camera extrinsic to avoid overfitting the decoder. 
We train the detector using an Adam optimizer with a one-cycle learning rate scheduler and a maximum learning rate of $10^{-3}$. Depending on the size of the encoder, a batch size of $1-2$ is employed. To stabilize training, we accumulate gradients over multiple batches before updating the weights to have an adequate batch size of $16$. The encoder and decoder network are initialized with weights pre-trained on \textit{ImageNet-1K}.
We run all experiments on RTX 3090 GPU.

\nparagraph{Temporal Caching}
Our method incorporates the \gls{bev} feature of the previous frame. A challenge is to have the previous feature cached while still achieving a high variation of batch samples, as the gradients are updated in order of the sequence. For testing, we can resort to a batch size of one to always have the previous feature computed, but for training, this would harm the performance due to a small batch size or slight sample variation. Thus, we build a custom sampler that composes batches according to the accumulated batch size in a semi-sequential order.

\begin{figure*}[t]
    \centering
    \begin{subfigure}[b]{0.33\textwidth}
      \includegraphics[width=\textwidth]{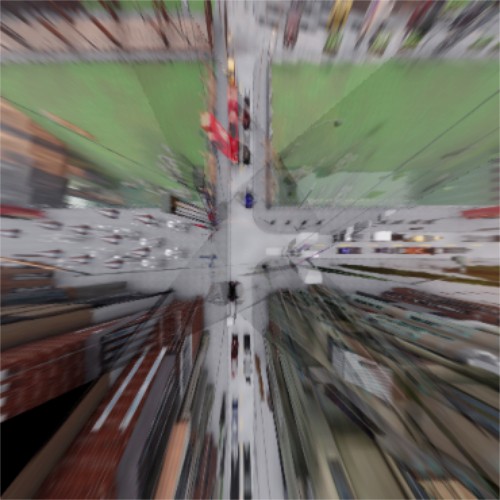}
      \caption{Projected \gls{bev}}\label{fig:quali_bev}
    \end{subfigure}
    \hfill
    \begin{subfigure}[b]{0.33\textwidth}
      \includegraphics[width=\textwidth]{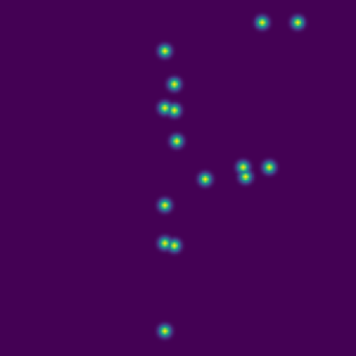}
      \caption{Ground Truth}\label{fig:quali_gt}
    \end{subfigure}
    \hfill
    \begin{subfigure}[b]{0.33\textwidth}
      \includegraphics[width=\textwidth]{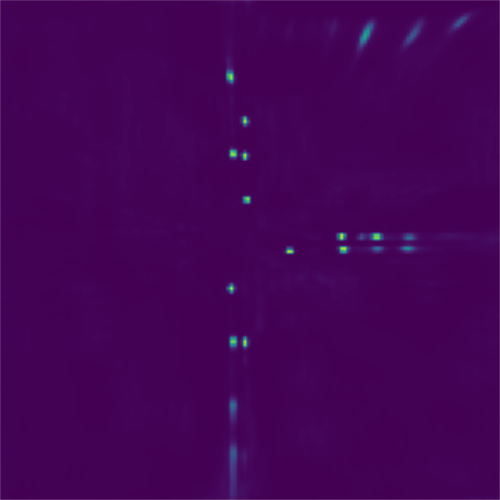}
      \caption{Prediction}\label{fig:quali_pred}
    \end{subfigure}
	\caption{\textbf{Qualitative Results.} Detection example shown on Synthehicle. (a) shows the input images projected to the BEV space, (b) shows the ground truth heatmap of all vehicles, and (c) our prediction with bilinear sampling.}
	\label{fig:quali}
\end{figure*}

\subsection{Results}
\nparagraph{Pedestrian} We report the detection results on both pedestrian datasets in \cref{tab:detection-sota}. Compared to previous works, our approach can improve the state-of-the-art further. Only the two-stage approach MVTT \cite{lee2023multi} is better in MODA on Wildtrack. Our approach dominates all other metrics. The overall high values indicate that the results start to saturate. The results on MultiviewX improve with larger margins compared to Wildtrack. These results may indicate that the labeling accuracy on Wildtrack limits us as they annotated using perspective transform \cite{chavdarova2018wildtrack} and underlines the need for more challenging datasets.

All recent related work used the perspective projection as the lifting method, and our approach can beat those approaches due to the additional temporal information. Overall, the parameter-free lifting methods, bilinear sampling, show competitive results for parameterized depth splatting. This observation aligns with the expectation that the depth splatting method can provide dense features throughout the observed space for this small area. The deformable attention method could perform better on our task, even though \cite{harley2022simple} showed it to be the most robust lifting method for autonomous driving tasks. Compared to the other methods, we observed strong overfitting effects during training, indicating that this method is not viable for small datasets like the one tested here. 

We compare the tracking results in \cref{tab:track-sota}. Our approach improves the \gls{sota} on both datasets. Our work can further improve the tracking quality compared to the three most recent approaches. It shows that tracking in \gls{bev} is currently the most potent approach, as EarlyBird  \cite{teepe2023earlybird}, MVFlow \cite{engilber2023multi}, ReST \cite{cheng2023rest}, and our approach follows this idea. EarlyBird  \cite{teepe2023earlybird} is most similar to our approach as it is also a one-shot tracker. However, it uses an appearance-based association and only the information of a single frame. The strength of our association method is mainly reflected in the IDF1 score, and the significant gain shows the advantages of our early-fusion tracker, which can combine appearance and motion cues.
Overall, we can observe similar trends to those of the detection task as the improvements stagnate on Wildtrack.   
\begin{table}[t]
    \centering
    \resizebox{\linewidth}{!}{
    \begin{tabular}{lccccc}
    \toprule
    & \multicolumn{2}{c}{Detection} & \multicolumn{2}{c}{Tracking} \\\cmidrule(lr){2-3}\cmidrule(lr){4-5}
    & MODA & MODP & IDF1 & MOTA  \\
    \midrule
    Baseline (Bilin. Sampl.)& 95.4 & 89.8 & 81.5 & 90.0 \\
    \midrule
    + History Fusion        & 96.5 & 75.0 & 83.8 &  87.9 \\
    + Motion Prediction     & " & " & 85.6 & 92.4 \\
    \bottomrule
    \end{tabular}
    }
    \caption{\textbf{Temporal Ablation.} Evaluation of temporal aggregation components introduced by our approach compared to the baseline on the MultiviewX dataset.}
    \label{tab:model-ablation}
\end{table}

\nparagraph{Vehicle Results}
In \cref{tab:synthehicle}, we report the results on Synthehicle \cite{synthehicle2023herzog} with the two lifting methods. The main advantage of evaluation on Synthehicle is that we can evaluate scenes known during training (scene-specific) and new scenes (cross-scene). The much more complex dataset shows much lower tracking scores. The parameterized depth splatting lifting method outperforms the parameter-free 
bilinear sampling by a significant margin. Both approaches do not show robust scene generalization capabilities, mainly in the detection quality, as indicated by the low MOTA score.  

\nparagraph{Temporal Aggregation}
The ability to model temporal information is crucial for a tracking model. In \cref{tab:model-ablation}, we ablate the aggregation components of our model. The baseline is our proposed method without access to the history and motion prediction. With additional history frames, the detection accuracy increases, but the precision decreases. This obeservation indicates that the model can detect more pedestrians with history frames, but the location precision decreases due to ambiguity introduced with the history frames. The additional motion prediction only affects the tracking results, and the results improve significantly from our prediction. 

\subsection{Qualitative Results}
In \cref{fig:quali}, we show an example from the Synthehicle validation set. First, in \ref{fig:quali_bev}, we projected all camera views perspectively to the ground plane to give an approximation of the \gls{bev} of the scene. Thus, it also approximates how the image features after the encoder is projected. The image features, or pixels, are stretched further on the outer parts of the scene. In \ref{fig:quali_gt}, we show all objects in the ground truth. Compared to the prediction (\ref{fig:quali_pred}) of our approach with the bilinear lifting method, we show promising results in the center of the scene. The strength of the prediction declines on the outer parts of the scene. The predictions also show vehicles in areas not highlighted in the ground truth but visible in the projected BEV-View. Synthehicle builds the label based on 2D views and thus might miss detections in 3D.

\section{Conclusion}
Our paper gives an extensive overview of different lifting strategies for \gls{mtmc} task. Combined with a motion-based tracking approach, we show \gls{sota} results on two pedestrian datasets. Our results show that the results are saturating on Wildtrack and MultiviewX, requiring new datasets. Thus, we extended our evaluation to a roadside perception dataset. This dataset allowed for a new challenge in this area: scene generalization. However, all datasets considered still focus on 2D detections, and our results show that the field needs new 3D-first datasets as a standard benchmark.
We are confident that our approach inspires a new \gls{mtmc} dataset and new one-shot multi-view detection and tracking approaches.

{
    \small
    \bibliographystyle{ieeenat_fullname}
    \bibliography{main}
}

\end{document}